# DEEP LEARNING FOR FACE RECOGNITION: A CRITICAL ANALYSIS

**ANDREW JASON SHEPLEY**

Charles Darwin University,
Ellengowan Drive, Casuarina, Darwin, Australia

Email: ashepley@myune.edu.au

**ABSTRACT**

*Face recognition is a rapidly developing and widely applied aspect of biometric technologies. Its applications are broad, ranging from law enforcement to consumer applications, and industry efficiency and monitoring solutions. The recent advent of affordable, powerful GPUs and the creation of huge face databases has drawn research focus primarily on the development of increasingly deep neural networks designed for all aspects of face recognition tasks, ranging from detection and preprocessing to feature representation and classification in verification and identification solutions. However, despite these improvements, real-time, accurate face recognition is still a challenge, primarily due to the high computational cost associated with the use of Deep Convolutions Neural Networks (DCNN), and the need to balance accuracy requirements with time and resource constraints. Other significant issues affecting face recognition relate to occlusion, illumination and pose invariance, which causes a notable decline in accuracy in both traditional handcrafted solutions and deep neural networks. This survey will provide a critical analysis and comparison of modern state of the art methodologies, their benefits, and their limitations. It provides a comprehensive coverage of both deep and shallow solutions, as they stand today, and highlight areas requiring future development and improvement. This review is aimed at facilitating research into novel approaches, and further development of current methodologies by scientists and engineers, whilst imparting an informative and analytical perspective on currently available solutions to end users in industry, government and consumer contexts.*

I. **INTRODUCTION**

Biometric recognition software plays an increasingly significant role in modern security, administration and business systems. Biometrics include fingerprint, retinal scanning, voice identification and facial recognition. Facial recognition has attracted particular interest, as it provides a discreet, non-intrusive means of detection, identification and verification, without the need for the subject's knowledge or consent. It is now commonplace in applications such as airport security, and has been widely embraced by law enforcement agencies, due to the improving accuracy and deployability of systems, and the growing size of face databases. A recent example of its success occurred in China; face recognition technologies were successfully used to identify and track a wanted fugitive at a concert attended by over 60000 people, resulting in his arrest [2]. As a consequence of the proven ability of deep neural network based systems to outperform human performance in face verification tasks, the international government sector is projected to be the largest user of face biometrics systems, including face recognition [3], over the next 10 years. Face recognition technologies are also at the forefront of a wide range of consumer applications and devices, from user verification tasks enabling account access, to digital camera applications, and social media tagging. Consumers and industry require rapid, affordable and efficient applications, to meet demand in business, employment and education functions, which extend to employee monitoring, roll call and security, reducing administrative costs, and procedural efficiency. Despite significant progress in recent years, and widespread usage, many shortcomings still exist. This survey will provide a comprehensive perspective on the current state of face detection, verification and identification technologies, highlighting limitations which must be rectified in order to progress to efficient, dynamic and versatile systems capable of meeting the needs of modern day usage.

Since the 1990s, significant progress has been made in the realm of face detection and recognition [4]. Current research in both face detection and recognition algorithms is focused on Deep Convolutional Neural Networks (DCNN), which have demonstrated impressive accuracy on highly challenging databases such as the WIDER FACE dataset [5] and the MegaFace Challenge [6], as well as on older databases such as Labeled Faces in the Wild (LFW) [7]. Rapid advancements have been triggered due to the increasing affordability of powerful GPUs [8], and improvements in CNN architecture design which is focused on real world applications [9]. Furthermore, large annotated datasets, and a better understanding of the non-linear mapping between input images and class labels has also contributed to the increase in research interest in DCNNs. DCNNs are very effective due to their strong ability to learn non-linear features, however they are inhibited by intensive convolution, and non-linear operations, which result in high computational cost [10]. Nevertheless, DCCNs are predicted to encompass future research and industry application, and are currently being deployed by large corporations such as Google, Facebook, and Microsoft [11].

No comprehensive survey of face detection, recognition and verification methods has been conducted since 2003, although several quality reviews have been conducted on specific aspects of facial recognition solutions. However, they lack complete coverage of all existing research areas and processes necessary to the development of effective solutions. To address this shortage, this paper will review all relevant literature for the period from 2003-2018 focusing on the contribution of deep neural networks in drastically improving accuracy. Furthermore, it will provide a comprehensive analysis of the state-of-the-art approaches in face detection, verification and identification, alongside a thorough coverage of the most recently developed databases and benchmarks, and preprocessing methods. Specific applications, such as video-based face recognition, and infrared recognition technologies will also be considered. Throughout the review, shortcomings and limitations are highlighted, and indicators of areas requiring future research and improvement are emphasized. Causes of inaccuracy and computational bottlenecks are explored, and recently proposed solutions evaluated in terms of real life applications. Furthermore, a comparative analysis of state of the art methods will be provided, as well as a brief overview of traditionally used methods which have recently been outperformed, but which provide an alternative to computationally expensive deep methods. In particular, this survey will identify areas of improvement in processing time, efficiency, cost and accuracy and analyse the means by which effectiveness is measured. Our contributions are outlined as follows;

- Provide visually appealing comparative analysis of modern face recognition systems and databases using tables, graphs and figures revealing performance and key features,
- Offer a critical analysis of the benefits and limitations of the state of the art methods, and a broad range of solutions designed to address shortcomings,
- Summarise the role of current research in the context of traditional methodologies, outlining the development of technologies over time,
- Highlight major shortcomings, and key areas requiring improvements in light of the latest research undertaken in specific areas of facial recognition.

## II. BRIEF CONTEXT

Since the seminal work of [12] introduced eigenfaces to the computer vision community, and [13] proposed the use of Haar features for face detection, there has been a significant increase in interest in facial recognition technologies. Technical progress in this field can be separated into four major areas of research interest, as shown in *Figure 1*. Holistic approaches employ distributional concepts such as manifold [[14], [15], [16]] and sparse representation [[17], [18]], and linear subspace [[19], [20]] to create low-dimensional representations. These approaches were however limited by the uncontrollable variations in facial appearance, which naturally deviate from distribution assumptions. This limitation gave rise to the use of local feature-based techniques such as Gabor wavelets [[21], [22] [23]], and LBP [24], which provided greater robustness due greater invariance to transformations and environmental effects. These methods were limited by lack of distinctiveness, a problem which was partially rectified by the use of shallow learning-based local descriptors [[25], [26], [27]] which improved compactness and introduced automated encoding. However, these shallow methods failed to achieve optimal performance due to their inability to accurately represent the complex non-linear variations in facial appearance.

| | | | |
|---|---|---|---|
| Deep Learning | **ArcFace** *2018* 99.83% (LFW) **SphereFace** *2017* 99.42% (LFW) **FaceNet** *2015* 99.63% (LFW) | | Current state of the art on LFW and MegaFace Challenge. Although identification rate on the latter is still low (82.55% on SphereFace), verification rate is close to 100%. Significant improvements in handling expression, pose, illumination and occlusion. |
| | | | High computational cost due to extensive use of GPUs and very deep network architectures. Issues of poor annotation and noise, together with image quality affect performance |
| | **DeepFace** *2014* 97.35% (LFW) | | Surpassed human verification accuracy in unconstrained settings for the first time. Commenced movement to focus research on deep learning methods such as CNNs |
| | | | High computational cost and suffers from loss of accuracy in situations involving spoofing, cross-pose, cross-age, low-resolution and make-up |
| Shallow | **PCANet** (*2015*) 86.28% (LFW) **LE (Learning-based Descriptor)** (*2010*) 84.45% (LFW) | | Improved distinctiveness and compactness of codebook. |
| | | | Representation not robust to complex non-linear nature of face |
| Local Handcraft | **Local Binary Patterns** *2004* 66-79% (FERET) **Gabor Wavelets** *2002* >70% (LFW) | | Robust to illumination and expression Removed the need for manual annotation |
| | | | Manually designing optimal encoding method and codebook is very difficult Susceptible to surface issues such as blurring. Results in uneven distribution which reduces informativity and compactness |
| | **Haar Features** *2001* 93.9% (Detection on MIT-CMU test set) | | Provided method of detecting faces efficiently and effectively. Pioneered boosting based detection methods. |
| | | | Sensitive to illumination, pose, image quality |
| Holistic | **EigenFaces** *1991* 60.02% (LFW) | | Simple, efficient method of recognizing faces in constrained environments. |
| | | | Relatively ineffective in face recognition in unconstrained conditions due to lack of robustness to lighting, pose, expression and image quality changes |

*Figure 1: Timeline of developments in facial feature representations and face verification accuracy*

A major milestone in the development of facial recognition techniques was achieved by the introduction of highly accurate deep learning methods such as DeepFace [28] and DeepID [29]. For the first time, face verification in unconstrained settings was achieved with accuracy surpassing human ability. This development was only allowed for by the advent of significant improvements in hardware, such as high capacity GPUs. Since then, the majority of research has focused on the development of deep learning-based methods which attempt to model the human brain, via high-level abstraction achieved using a concurrence of non-linear filters resulting in feature invariance. The majority of these methods rely on increasingly deep CNNs, with an emphasis on promoting sparsity and selectivity. Other deep learning methods.

### III. RELATED WORKS

Facial recognition and verification is an area of high research interest due to its broad span of applications, the available scope for improvement in accuracy and computational speed due to innovation in hardware and increasingly large and accessible databases. Accordingly, literature reviews have been conducted periodically to cover these changes. However, due to the vast range of face recognition methods employed, most reviews focus on a particular issue or set of problems, rather than addressing the entire range of dominant methods. For example, several recent surveys have specifically addressed a range of methodologies which have attempted to achieve rotation invariant face recognition (Oscos, Khoshgoftaar [30]), pose-invariant face recognition [31], skin-based face detection in various conditions [32] and low-resolution face recognition [33]. Other publications have reviewed face recognition techniques from various perspectives, for example, only 2D methods were reviewed in [34] whilst several papers, including [35] and [36] examined face recognition in terms of presentation attack detection and anti-spoofing technologies. However, these surveys lack a comprehensive coverage of all currently relevant facial recognition methodologies, and often do not refer to the most current databases and benchmarks, such as the MegaFace Challenge benchmark.

Another disadvantage reflected in several recent surveys including [37], [38], [39] and [40] is the excessive focus on local handcraft descriptors and shallow learning methods. In contrast, this review presents a strong emphasis on the current research trends, particularly in the realm of deep learning methods, such as the usage of DCNNs, which currently produce state of the art standards in face detection, recognition and verification tasks. It must also be noted that various quality reviews have been conducted on aspects of face recognition, such as face detection alone, [[41], [42]], or identification and verification tasks [43]. In contrast, this survey will attempt to provide a clear comparison and analysis of the advantages and disadvantages of a comprehensive range of techniques in all areas of face recognition including detection, feature extraction and classification. It will also offer an overview of which methodologies are most suited to a range of applications whilst highlighting the areas within each area of research which could be further improved. Attention is also given to the latest databases and benchmarks used to measure the accuracy and scope of face recognition systems, noting whether they are publicly available or not.

## IV. DATABASES

All facial recognition and detection systems require the use face datasets for training and testing purposes. In particular, the accuracy of CNNs is highly dependent on large training datasets [44]. For example, the development of very large datasets such as ImageNet [45], which contains over 14 million images, has allowed the development of accurate deep learning object detection systems [11]. More specifically, face detection and recognition datasets developed alongside benchmarks such as the MegaFace Challenge [46], a subset of which is shown in *Figure 2*, the Face Detection Dataset and Benchmark (FDDB) dataset [47] and the Labeled Faces in the Wild (LFW) dataset [48] provide a means to test and rank face detection, verification and recognition systems using real-life, highly challenging images in unconstrained settings. Notable and widely used datasets are listed in *Table 1*, along with information regarding their intended usage, size and the number of identities they contain.

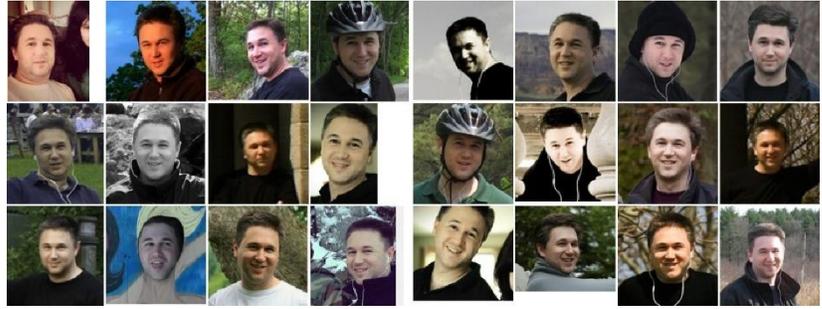

*Figure 2: Sample subset of the MegaFace Challenge dataset*

Upon analysis of the results attained by face verification and identification algorithms tested on small datasets such as the LFW dataset, one may be led to believe there remains little scope for improvement. This is far from true: when tested on millions of images, algorithms achieving impressive results on smaller testing sets produce far from ideal accuracies [46]. The MegaFace Challenge was created in response to the saturation of small datasets and benchmarks, providing a large-scale public database and benchmark which requires all algorithms to be trained on the same data and tested on millions of images, allowing fair comparison of algorithms without the bias of private dataset usage. This addresses the problem of lack of reproducibility of results [49] caused by the usage of private databases for training by state of the art CNN methods [50]. Although a shortage of cross-age identity sets is one limitation of the MegaFace dataset, results thus far have indicated there is ample scope for algorithm improvement, with the highest identification and verification accuracies attained by the state of the art method ArcFace [49] reaching 82.55%, and 98.33% respectively. Similarly, the MS-Celeb-1M database was created to provide both training and testing data, to enable the comparison of face recognition techniques by use of a fixed benchmark. However, despite the benefits conferred by their size, both MegaFace and MS-Celeb-1M are disadvantaged by annotation issues [51] and long tail distributions [52].

| | Database | Website | Features | Application |
|---|---|---|---|---|
| **Public** | **MegaFace** [6] | http://megaface.cs.washington.edu/index.html | 4,700,000 images 672,000 identities | *Large database and benchmark suited for CNN comparison.* |
| | **WIDER FACE** [5] | http://mmlab.ie.cuhk.edu.hk/projects/WIDERFace/ | 32,203 images containing 393,703 faces | *Face detection with large illumination, expression, makeup, occlusion, scale and pose variations* |
| | **Labelled Faces in the Wild (LFW)** [7] | http://vis-www.cs.umass.edu/lfw/ | 13,233 images 5749 identities | *Benchmark for automatic still image face verification* |
| | **VGG-Face** [48] | http://www.robots.ox.ac.uk/~vgg/data/vgg_face/ | 2,600 000 images. 2,622 identities. | *Celebrity images in unconstrained conditions* |
| | **VGG2-Face** [53] | http://www.robots.ox.ac.uk/~vgg/data/vgg_face/ | 3,310,000 images 9131 identities | *Downloaded from Google Image Search* |
| | **MS-Celeb-1M** [54] | https://www.msceleb.org/ | 10,000,000 images 100,000 identities | *Celebrity images with associated entity keys* |
| | **FaceScrub** [55] | http://vintage.winklerbros.net/facescrub.html | 107,818 images 530 identities | *Celebrity images in unconstrained conditions* |
| | **FERET** [56] | http://www.cs.cmu.edu/afs/cs/project/PIE/MultiPie/Multi-Pie/Home.html | 14,126 images. 1199 identities | *Used for comparing effectiveness of recognition algorithms* |
| | **CASIA-WebFace** [57] | http://www.cbsr.ia.ac.cn/english/CASIA-WebFace-Database.html | 494,414 images. 10,575 identities | *Large database with many identities* |
| | **MAFA dataset** [58] | http://www.escience.cn/people/geshiming/mafa.html | 30,811 images containing 35,806 faces | *Face detection focusing on masked or occluded faces* |
| | **CMU Multi-PIE** [59] | http://www.cs.cmu.edu/afs/cs/project/PIE/MultiPie/Multi-Pie/Home.html | 750,000 images. 337 identities | *Varying illumination and expression* |
| | **MORPH Database** [60] | http://www.faceaginggroup.com/morph/ | 1724 images. 515 identities (1962-1998) More recent album: 55,134 images. 13,000 individuals | *Large longitudinal face database showing effects of aging* |
| | **CACD** [61] | http://bcsiriuschen.github.io/CARC/ | 163,446 images. 2000 identities | *Focuses on age related issues* |
| | **YouTube Faces (YTF)** [62] | https://www.cs.tau.ac.il/~wolf/ytfaces/ | 3425 videos. 1595 identities. | *Benchmark for unconstrained video-based face verification* |
| | **SCface** [63] | http://www.scface.org/ | 4160 images. 130 identities | *Video surveillance camera footage* |
| | **UMDFaces** [64] | http://www.umdfaces.io/ | 367,888 images 8,277 identities & 3,700,000 video | *Clean, highly annotated celebrity/public figure images obtained using GoogleScraper* |

| | | | frames of 3100 identities | |
|---|---|---|---|---|
| | **Face Detection Dataset and Benchmark (FDDB)** [47] | http://vis-www.cs.umass.edu/fddb/ | 2845 images 5171 faces | *Face detection for images containing 1 or more faces* |
| | **PolyU near-infrared face database (PolyU-NIRFD)** [65] | http://www4.comp.polyu.edu.hk/~biometrics/polyudb_face.htm | 35,000 samples. 350 identities | *Used for NIR face recognition in controlled and uncontrolled conditions* |
| | **3D Mask Attack Database (3DMAD)** [66] | https://www.idiap.ch/dataset/3dmad | 76,500 images. 17 identities. | *Recorded using Kinect. Aimed at preventing 2D face spoofing attacks.* |
| **Private** | **Social Face Classification (SFC)** [28] | Unavailable | 4,400,000 images. 4030 identities | *Facebook internal dataset* |
| | **FaceNet** [67] | Unavailable | 100M-200M images, 8,000,000 identities | *Google dataset of face thumbnails* |
| | **Face Detection at Night in Surveillance (FDNS)** [68] | Unavailable | 741 images. 1856 faces. | *Surveillance footage at nighttime for detection* |
| | **CrowdFaceDB** [69] | Unavailable | 385 videos. 257 identities | *Crowd video-based detection and recognition* |
| | **Bosphorus Database** [70] | http://bosphorus.ee.boun.edu.tr/Home.aspx conditionally available | 4666 images. 105 identities | *3D and 2D based – contains occlusions, expressions and poses* |
| | **KinectFaceDB** [71] | http://rgb-d.eurecom.fr/ conditionally available | 468 images. 52 identities | *Multimodal – 2D, 2.5D and 3D and video data using Kinect sensor* |
| | **Photoface Database** [72] | http://www1.uwe.ac.uk/et/mvl/projects.aspx#PPDB | 7356 images. 261 identities | *3D and 2D – contains natural real-life expressions and poses* |

*Table 1: Summary and comparison of the main features and focuses of publicly available face detection and recognition datasets, and notable private datasets*

The LFW dataset and benchmark is often referred to as the de facto benchmark for automatic still image face verification. It is a relatively small database which has been varied to include additional, specific databases. These variations are very useful for developers when addressing problematic aspects or issues associated with face recognition. Another small database is the FERET benchmark and database, which was developed to provide a means to directly compare different face recognition algorithms, identify state of the art methodologies, identify promising approaches, and highlight areas requiring future research to further the development of face recognition technologies. [73]. These benchmarks have thus far proven very useful in this purpose but fail to account for the significant data needs of currently used CNN approaches. For this reason, significantly larger databases and associated benchmarks including MegaFace, are being used for comparison of deep learning face recognition approaches.

## V. FACE DETECTION

Face detection is a fundamental step in facial recognition and verification [74]. It also extends to a broad range of other applications including facial expression recognition [75], face tracking for surveillance purposes [76], digital tagging on social media platforms [77] and consumer applications in digital technologies, such as auto-focusing ability in phone cameras [78]. This survey will examine facial detection methods as applied to facial recognition and verification. Historically, the greatest obstacle faced by face detection algorithms was the ability to achieve high accuracy in uncontrolled conditions. Consequently, their usability in real life applications was limited [41]. However, since the development of the Viola Jones boosting based face detection method [13], face detection in real life settings has become commonplace. Significant progress has since been made by researchers in this area [41] due to the development of powerful feature extraction techniques including Scale Invariant Feature Transform (SIFT) [79], Histograms of oriented Gradients (HoGs) [80], Local Binary Patterns (LBPs) [81] and methods such as Integral Channel Features (ICF) [82]. For a recent and comprehensive review of these traditional face detection methodologies, readers are referred to [83]. This review will alternatively focus on more recently proposed deep learning methods, which were developed in response to the limitations of HoG and Haar wavelet features in capturing salient facial information under unconstrained conditions which include large variations in resolution, illumination, pose, expression, and color. Essentially, it is the limitations of these feature representations which have thus far limited the ability of classifiers to perform to the best of their ability [43]. Furthermore, due to the significant increase in availability of large databases, DCNNs generally demonstrate higher performance in object and face detection tasks, as demonstrated by [11], [84] and [85].

*Deep Learning: Face Detection*

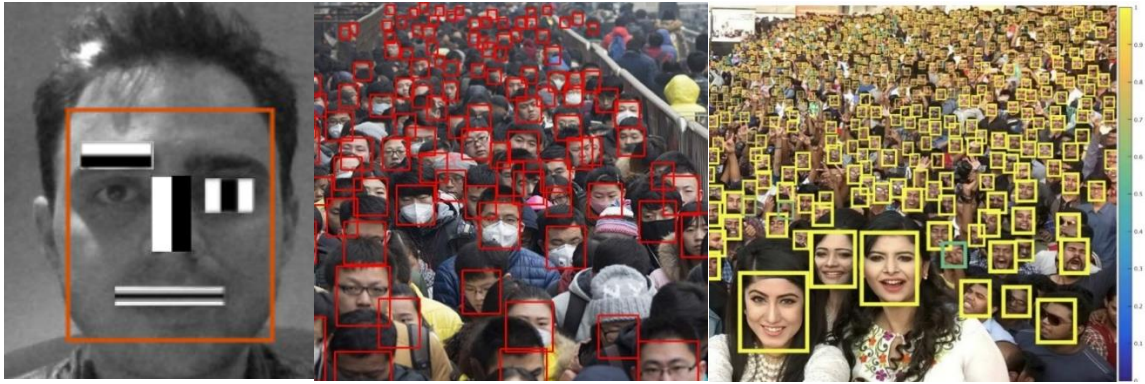

*Figure 3: From Haar features to tiny, and highly occluded faces. The use of CNNs in face detection has significantly improved accuracy*

Recently, the creation of large annotated databases such as the MegaFace Challenge, LFW and WIDER FACE has encouraged the development of highly discriminative, state of the art deep learning face detection. Consequently, DCNNs now perform significantly better in object and face detection tasks, as demonstrated by [11], [84] and [85]. DCNN face detection methods can be categorized as region based or sliding window approaches. The region-based approach uses an object proposal generator such as Selective Search [86] to generate a pool of regions which may include one or more faces. These proposals are then input into a DCNN which classifies them as either including a face or not, returning the precise bounding box coordinates of faces in a given image with minimal background inclusion [43]. Most current methods, including HyperFace [87] and All-in-One-Face [88] use this approach. More efficient region-based CNNs (R-CNN) [89] use a DCNN to generate proposals, perform bounding box regression and classification. [90] improved upon R-CNN by using a combination of feature concatenation, multi-scale training improving scale invariance, and hard negative mining, to reduce false positive rates, achieving state of the art recall and accuracy. This method suffers from high computational cost, thus requires improvements in efficiency and scalability to be deployable for real-time face detection. Other methods which aim to improve upon state of the art region based methods include [91], which developed a method to reduce redundant region proposals, and [92], who proposed a very lightweight Single-State Headless face detector which achieved state of the art accuracy by detecting faces directly from the early convolutional layers within the classification network. However, despite impressive results, region-based deep face detectors are computationally expensive due to the requirement of proposal generation.

An alternative and far more efficient approach to face detection is the sliding window-based method, which computes accurate bounding box coordinates at each location in a feature map of a specific scale, using a convolution operation. Scale invariance is achieved by generating an image pyramid containing multiple scales. One such method is [97], which proposed a single-shot detector which uses the inbuilt pyramidal DCNN cascade architecture to rapidly eliminate background regions at low resolutions, allowing only challenging regions to be processed at high resolutions. A single forward pass is sufficient to obtain detections, thus reducing computation time. [98] also achieves superior results by use of a novel hard sample mining strategy together with a deep cascaded multitask framework which leverages off the correlation between detection and alignment to improve performance. Another sliding window-based method is [99] which developed DP2MFD, a deformable parts model integrated with deep pyramid features, wherein the face is defined as a collection of parts which are trained alongside the global face, to achieve scale invariant, state of the art detection. Furthermore, [100] attempted to address the issue of multi-view face detection by proposing a minimally complex Deep Dense Face Detector (DDFD) without the need for pose or landmark annotation. It achieved similar results to highly complex methods but is limited due to inadequate sampling strategies and the need to improve data augmentation. Faceness [101] claimed to achieve effective face detection even in cases where over 50% of a given facial region was affected by occlusion. It also claimed to overcome significant pose variation, with the added benefit of accept arbitrary images of varying scale. This was achieved using a set of attribute aware deep networks which were pre-trained with generic objects, followed by refinement using specific part-level binary attributes. However, the authors acknowledged that improvements in speed were possible, noting the benefits of integrating model compression techniques and approximation of non-linear filtering with low-rank expressions [102]. Another recently proposed method is ScaleFace: [93], which designed a simplified multi-network CNN approach capable of detecting faces at a very wide range of scales, by using a specialized set of DCNNs with varying structures, without the use of a traditional image pyramid input. Finally, [94] evaluated the significant impact of contextual information on the detection of very small faces, subsequently developed massively-large receptive field based templates used to train separate detectors for different scales, improving the state of the art results on WIDER FACE from 29-64% to 81% . *Figure 3* shows illustrates the significant progress achieved by face detection systems when detecting naturally occluded faces, handling significant discrepancies in scale.

| Face Detection Method | WIDER FACE (hard) |
|---|---|
| **ScaleFaces** [93] | 76.4% |
| **HR** [94] | 81.9% |
| **SSH** [92] | 84.4% |
| **S3FD** [95] | 85.8% |
| **FAN** [96] | 88.5% |

*Table 2: Top 5 face detection methods*

Despite the increasing accuracy and speed of face detection systems, the two greatest challenges remain somewhat unresolved. Face detectors are required to cope with large and complex variations in facial changes, and effectively distinguish between faces and non-faces in unconstrained conditions. Furthermore, the large variation in face position and size within a large search space presents challenges which reduce efficiency [103]. This calls for a trade-off between high accuracy and computational efficiency. One benefit of less accurate Viola Jones inspired cascade-based face detectors over CNN methods is their efficiency. Thus the greatest requirement in the current field of research is the development of more efficient CNN face detection techniques. [96] partly addressed this issue, achieving the current state of the art accuracy rate of 88.5% on the hard WIDER FACE test set by developing the Face Attention Network (FAN), a novel face detector designed to improve recall in cases of occlusion without impacting on computation speed. This was achieved by using an anchor-level attention to enhance facial features within a face region, together with random crop data augmentation to tackle occlusion and tiny faces. A comparison of the five highest accuracy face detectors as measured on the WIDER FACE benchmark is provided in *Table 2*.

## VI. FEATURE EXTRACTION

Feature extraction usually occurs immediately after face detection and can be considered as one of the most important stages in face recognition systems, as their effectiveness is dependent upon the quality of the extracted features. This is because facial landmarks and fiducial points identified by a given network determine how accurately features are represented. Traditional fiducial point locators are model-based, whilst many recent methods are cascaded regression based [43]. Lately, key improvements have been made with the development of deep dual pathway methods [1], and other confidence map based solutions, such as [104] and [105]. Traditional model-based fiducial point methodologies include Active Shape Model (ASM), which suffers from low accuracy, partially rectified by the work of [106], Active Appearance Model (AAM) [107], and Constrained Local Models (CLM). CLMs are generally outperformed by cascaded regression [108], models due to the latter's inherent inability to mode the complex variation of local feature appearances. It must be noted however, that highly effective methods based on CLMs have been developed. For example [109] is based on CLMs but takes advantage of the neural network architecture, proposing a Convolutional Experts Network (CEN) and Convolutional Experts Constrained Local Model

(CE-CLM) which uses CLM as local detector, achieving very competitive results particularly on profile images.

Face recognition differs to object recognition in that it involves alignment before extraction [50]. This is reflected in the differences between CNNs used for face recognition and those used for object recognition. An increase in data availability has resulted in development of learning-based methods as opposed to engineered features due to their inherent ability to discover and optimize features specific to a task. Consequently, learning methods have outperformed engineered features [28]. In CNNs, a fiducial point detector is employed to localize important facial features such as eye centers, mouth corners and nose tip. Once these landmarks have been identified, the face is aligned and according to normalized canonical coordinates [110]. Subsequently, a feature descriptor is extracted, encoding identity information. Similarity scores are then calculated using these face representations. If this score is below a set threshold, the faces are classified as belonging to the same identity. Model based DCNN methodologies employ training to learn a shape model, which is used to fit new faces during testing. However, these learned models are limited by their sensitivity to gradient descent optimization initialization and their lack the ability to represent complex facial variations in pose, expression and illumination [1]. Deep cascade regression based models, first proposed by [111], and improved by [112], rapidly outperformed shallow methods such as [113] and [114]. These methods learn a model which directly maps the appearance of the image to the target output. The effectiveness of these methods is highly dependent on the robustness of the local descriptors used. However, most cascaded regression methods involve independent regressor learning, which causes issues of cancelling regressor descent direction, thus inhibiting learning. To address these shortcomings, [115] proposed a combined convolutional recurrent network which allows training of an end-to-end system, in which the recurrent module facilitates joint optimization of regressors by assuming cascades are a non-linear, dynamic system. Thus, all information between cascade levels is used and shared between layers.

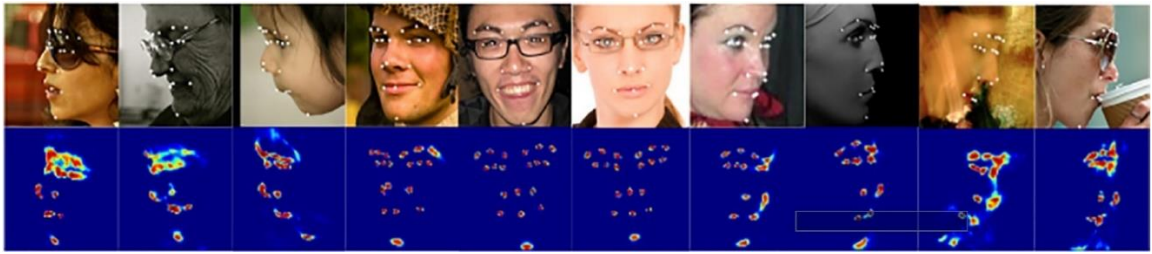

*Figure 4:* Fiducial point detection by the state-of-the-art method proposed by [1]

Subsequently, more representative deformable 3D models used by [116] and [117] to estimate facial poses and shape coefficients outperformed then state of the art shallow methods, including [118] and [119]. These deep methods are however disadvantaged by the need to re-initialize models when switching stages or networks, particularly in systems where local deep networks are used to localize fiducial points based on facial patches, as seen in [111]. [120] used cascade regression to predict a 3D to 2D projection matrix and base coefficients. The concept was further developed in [116] which approached face alignment from a 3D model fitting perspective, resulting in the development of a cascade of DCNN-based regressors which function to estimate 3D shape parameters, and the camera projection matrix. [121] addressed the problem of variations in pose by using a simple, generic 3D surface to approximate the shape of all input faces. This method is however limited by its reliance on the quality of detected landmarks which, if poor, can cause the appearance of undesirable artifacts. Other works include [122] which used a DCNN to fit a dense 3D face model to a given image, employing a Z-buffer to model depth data. Alternatively [123] developed Local Deep Descriptor Regression (LDDR) which provides a highly accurate means of localizing fiducial points using deep descriptors which are able to accurately describe every pixel in a given image. [124] further presented an iterative method for unconstrained fiducial point estimation and pose production by employing a novel CNN architecture dubbed Heatmap-CNN (H-CNN) which captures both global and local features by generating a probability value, indicating the presence of joint at a defined location. This allows accurate, state of the art key point detection without the use of 3D mapping.

Multitask learning (MTL) approaches integrate face detection and fiducial point estimation within the same process, allowing greater robustness, due to additional supervision. One effective example is [91] which proposed Supervised Transformer Network, a cascade CNN which uses a two stage process to predict face candidates and landmarks, followed by mapping landmarks to canonical positions to normalize face patterns, followed by validation, achieving previous state of the art results. [125] presents a semantic approach which uses a combination of a ConvNet with a 3D model to detect faces and their fiducial points in the wild, achieving competitive results. It must be noted that network design significantly affects performance [1]. The abovementioned systems face difficulty in pixel-level localization and classification tasks due to spatial-semantic uncertainty [126] caused by the failure by to retain adequate spatial resolution after pooling and convolutional layers are employed when using deep representation generated at the lower layers. To rectify this issue, [87] concatenated shallow-level convolutional layers to the latest convolutional layers prior to landmark regression, while [1] proposes a dual pathway model which forces shallow and deep network layers to maximize the likelihood of highly specific candidate region. Notably, [127] used aggregation of shallow and deep layers to generate more accurate score map predictions, in the field of pose estimation.

Alternative methods include [128] [104] and [105], which generate a confidence map for each landmark to indicate likelihood of landmark appearing at specific location in original image. Prediction is made by selecting the location with maximum response as shown by the confidence map. In comparison to cascade regression these methods are more effective, as they suppress false predictions caused by noisy regions, thus improving robustness, with greater accuracy in unconstrained conditions. [128] addresses the problem of reliance on high quality detection bounding boxes coordinates by proposing a Convolutional Aggregation of Local Evidence (CALE) which comprised of a CNN which performs facial part detection, mapping confidence scores for the location of each landmark within the first few layers. The score maps and CNN features are then aggregated by using joint regression in order to refine landmark location. CNN regression guides contextual learning when predicating the position of occluded landmarks in unconstrained conditions, thus increasing robustness. However, networks such as [105] and [129] which rely on [130] to find the location of facial landmarks, are disadvantaged by the low quality confidence map generated by DeconvNet. Other methods such as [104] minimize residual error in score maps by use of a stacked cascaded architecture which refines key-point predictions, however it is difficult to deploy on a small scale due to its heavy and largely redundant architecture. Furthermore, [131] proposed an Lapalcian-pyramid architecture that provides effective refinement of 2D score maps generated by lower layers, by supervising the adding back of higher level generated features using three softmax layers. [1] proposed a Globally Optimized Dual-Pathway (GoDP) deep architecture to rectify these issues. This method aims to identify target pixels by solving a cascaded pixel labeling problem without the use of high-level inference models or complex stacked architectures. High quality 2D score maps are generated without the use of stacked architecture, partially rectifying lack of spatial semantic information by discriminatively extracting it from the deep network, as shown by *Figure 4*, and developed a novel loss function which reduces false alarms. This method currently achieves state of the art results, outperforming cascaded regression-based models on complex face alignment databases.

Currently, the greatest shortcoming present in the realm of unconstrained face alignment and fiducial point detection is the lack of solution to the problem of aligning faces irrespective of pose variation, and the general reliance of systems on accurate face detection. The 300 Faces in the wild database [132] is generally used for comparison of fiducial point detection methods. This face dataset is limited, and thus one area of improvement could include the creation of a large-scale annotated dataset containing a broad range of unconstrained facial images specifically designed for use in face alignment and fiducial point detection applications. This would improve robustness across fiducial point detection generally, particularly with respect to pose and expression variations, low illumination and poor quality. With respect to network structures, deepening neural networks may capture more abstract information which may assist in detection, however it is still unclear which network layers contribute most significantly to local features relevant to fiducial point detection [43]. This is one area which may benefit from further research. Furthermore, the high computational cost associated with localizing fiducial points still remains a significant challenge in unconstrained conditions.

## VII. FACE IDENTIFICATION AND VERIFICATION

Subsequent to feature extraction, facial recognition is performed. Recognition can be categorized as either verification or identification. Modern face recognition systems using DCNNs involve deep feature extraction, and lastly, similarity comparison. More specifically, verification involves comparison of one-to-one similarity between a probe image and a gallery of a known identity, whilst identification determines one to many similarities to determine the identity of the probe. Both these processes require robust feature representation, and a discriminative classification model or similarity measure. Traditional methods used for feature representation include LBP, HoGs, and Fisher Vector. Relevant metric learning methods include cosine metric learning, Mahalanobis metric learning, and one-shot similarity kernel. Others include large margin nearest neighbor, Joint Bayesian and attribute-based classifiers. These methods are thoroughly reviewed by [41]. Thus, for the sake of relevance and context, we have only included a brief overview of the role these methods play in modern face recognition and have chosen to focus on the most recently developed state of the art methodologies, which largely rely on DCNNs.

The modern CNN framework was designed in 1990 by [133] when they developed a system known as LeNet-5 to classify handwritten digits by recognizing visual patterns from image pixels without the need for preprocessing. [134] first presented a neural network used for upright, frontal, grayscale face detection, which although primitive by today's standards, compared in accuracy with state-of-the-art methods at the time. Since then, research has accelerated significantly, leading to the development of highly sophisticated DCCNs capable of detection, recognition and verification with accuracy approaches that of humans. Although the development of CNNs was impeded by lack of computing power [135], recent hardware advances have allowed rapid improvement and a significant increase in CNN depth, and consequently, accuracy. One outstanding feature is an increase in depth, and width to allow for improved feature representation by improving non-linearity [135]. However, this leads to issues such as reduction in efficiency and overfitting [9]. This section will explore the various methods which have aimed to address these problems in the context of facial recognition, through an examination of general improvements in DCCN architecture and loss functions. CCNs are generally more suitable to object recognition than standard feedforward neural networks of similar size due to the use of fewer connections and parameters which facilitates training and efficiency, with only slight reduction in performance [11]. CNNs were designed specifically for classification of 2D images [136] due to their invariance to translation, rotation and scaling [137]. A CNN is comprised of a set of layers, including convolutional layers, which are a collection of filters with values known as weights, non-linear scalar operator layers, and down sampling layers, such as pooling. Activation values are the output of individual layers which are used as input in the next layer [138]. For a thorough overview of basic CNN components, readers are referred to [135].

The use of CNNs in facial recognition tasks is comprised of two essential steps; namely, training and inference. Training is a global optimization process [135] which involves learning of parameters via observation of huge datasets. Inference essentially involves the deployment of a trained CNN to classify observed data [138]. The training process involves minimization of the loss function to establish the most appropriate parameters, and determination of the number of layers required, the task performed by each layer, and networking between layers, where each layer is defined by weights, which control computation. CNN face recognition systems can be distinguished in three ways; the training data used to train the model, the network architecture and settings, and the loss function design [49]. DCNN's have the capacity to learn highly discriminative and invariant feature representations, if trained with very large datasets. Training is achieved using an activation function, loss function and optimization algorithm. The role of the loss function is to determine the error in the prediction. Different loss functions will output different error values for an identical prediction, and thus determine to a large extent the performance of the network. Loss function type depends on the type of problem, e.g. regression or classification. Minimization of the error is achieved using back propagation of the error to a previous layer, whereby the weights and bias are modified. Weights are learned and modified using an optimization function, such as stochastic gradient descent, which calculates the gradient of the loss function with respect to weights, then modifies weights to reduce the gradient of the loss function [138].

| Model/ Method | Training Dataset | Number of NNs | Loss function | Face verification | Face idenitification |
|---|---|---|---|---|---|
| DeepFace [28] | SFC | 3 | Cross-entropy loss | 97.35% | - |
| DeepFR [48] | VGG-Face | 1 | Triplet | 98.95% | - |
| CenterFace [139] | CASIA-WebFace, CACD, Celebrity+ [140] | 1 | Center Loss | 99.28% 76.72% (MF1) | 65.23% (MF1) |
| SphereFace [141] | CASIA-WebFace | 1 | Angular softmax | 99.47% 89.14% MF1 | 75.77% MF1 (small protocol) |
| DeepID2+ [142] | CelebFaces+, WDRef | 25 | - | 99.47% | - |
| DCFL [143] | CASIA-WebFace | 1 | Correlation loss | 99.55% | - |
| FaceNet [67] | FaceNet | 1 | Harmonic triplet loss | 99.63% 86.47% (MF1) | 70.49% (large protocol) |
| CosFace [144] | CASIA - WebFace | 1 | Large Margin Cosine | 99.73% 97.96% (MF1) | 79.54% (small protocol) 84.26% (large protocol) |
| **ArcFace (LResNet100E-IR) [49]** | **Refined MS-Celeb-1M, VGG2** | **1** | **Additive Angular Margin Loss** | **99.83% (LFW) 98.48 (MF1)** | **83.27% (MF1)** |

*Table 3: State-of-the-art and competitive face verification and identification methods. All verification results are recorded on the LFW dataset unless indicated. All identification results are obtained on the MegaFace Challenge 1 (MF1) dataset.*

Loss function modifications have been very popular as a means of improving accuracy of face recognition systems, thus many variations have been proposed lately. Softmax loss function and its variations [145] are commonly used as they promotes separation of features. However, it shows ineffectiveness when intra-variations are greater than inter-variations. As such novel loss functions such as the ArcFace [49] loss function have been proposed, and have shown greater effectiveness. The additive angular margin (ArcFace) loss function produces more accurate geometrical interpretation than previously used supervision signals, obtaining more discriminative deep features by maximizing the decision boundary in angular space premised on L2 normalized weights and features. It currently produces the start of the art face identification and verification results on both LFW and the MegaFace Challenge, as shown by *Table 3*. This followed significant progress in the development of multiplicative [141] and additive cosine margins [146] whch are added into the Softmax loss to enhance its discriminative power. These angular and cosine margin-based loss functions have shown improved performance over Euclidean-distance based loss functions due to the use of angular similarity and separability between learned features. Particularly, [141] proposed an angular loss function based on the Softmax loss which uses highly discriminative feature representation optimized for cosine distance and similarly metric, achieving prior state of the art results. [147] proposed a combination of a novel triplet loss function, and feature fusion across layers which achieved state of the art performance in video-based face recognition, while [139] proposed a loss which uses the centroid in each class as a regularization constraint within the softmax function within a residual neural network. [145] used Softmax loss regularized with a scaled L2 Norm constraint which was shown to optimize the angular margin between classes. The last stage in face recognition is similarity comparison, which occurs after training. This involves the conversion of test images to deep representations, similarity is calculated by use of L2 distance or cosine distance, after which methods such as nearest neighbor or threshold comparison are used to identify or verify faces. Other methods, including metric learning and sparse representation classifiers are also used to post-process deep features to improve accuracy and efficiency. It must however be noted that despite the high accuracy produced using these novel loss functions, they suffer excessive GPU memory consumption within the classification layer when handling large amounts of data. Additionally, the triplet and contrastive loss functions are disadvnateged by the difficult task of selecting effective training samples.

Notably, [28] proposed DeepFace, which uses a Siamese network architecture which employs the same CNN to obtain descriptors for pairs of faces which are then compared via Euclidean distance. This method uses metric learning during training to minimize difference between two images of the same identity, and maximise the distance between those of differing identities. Although this process achieved state of the art recognition, it was further improved in [148] by increasing the size of the training data set. DeepFace was further enhanced by REF 24-27 in [48]. Generally, the architecture of CNNs is determined on experience, on a trial and error basis. [136] proposed to rectify this by developing a fully automated Adaptive Convolution Neural Network (ACNN) to specifically address facial recognition. Its structure is created automatically based on performance and accuracy requirements. Based on simple network initialization, convergence is then used to determine whether or not expansion will occur depending on the allowable system average error and desired recognition rate. It improves upon an Incremental Convolutional Neural Network (ICNN) proposed by [149] as global expansion is controlled automatically, rather than artificially. This study aimed to achieve a desirable balance between training time and recognition rate without the need for performance comparison [136]. [28] also took an alternative approach as they did not use standard convolution layers, instead relying heavily on an extensive database of over 4 million faces to train a nine-layer deep feedforward neural network, and a 3D face model based alignment, to generate a face representation. This network employed several locally connected layers without weight sharing and over 120 million parameters. This method claimed to achieve close to human level accuracy performance on the LFW dataset. However, the DeepID frameworks [142, 150] were however the first to achieve state of the art verification results which outperformed human performance, with the added benefit of using a smaller dataset. These approaches involved learning of highly discriminative and informative features by using a collection of smaller, shallow networks, and deep convolutional networks, specific to local and global face patches.

Improving performance by increasing depth and width has drawbacks such as overfitting, which may lead to bottlenecks and needlessly increases computer resources, e.g. when a lot of weights eventuate with 0 values [9]. This can be solved by modelling biological networks in transitioning to thinly connected architectures rather than fully connected networks. [9] proposed a DCNN named Inception, designed based on the Hebbian principle, i.e. neurons that fire together wire together, and multi-scale processing to maintain a computational budget of 1.5 billion multiply-adds at inference time to ensure cost effective real-world usage, on large databases. It claimed to outperform state of the art object detection and image classification by focusing on improving the structure of CNN. A 22-layer deep model was created, using 1 x 1 convolutions as dimension reduction modules to remove computational bottlenecks allowing both depth and width of the networks to be increased without impeding performance. With a similar goal of reducing computational cost [138] highlighted the need for greater sparsity. Sparsity is defined by as the proportion of zero values in a given layer's activation and weight matrices. The goal of achieving sparsity resulted in the creation of the Sparse Convolutional Neural Network (SCNN) architecture which was designed to enhance computational efficiency and performance at inference by manipulating zero valued activations and weights, to minimize unnecessary data processing and storage [138]. This is achieved using the sparse planar-tiled input-stationary Cartesian product (PT-IS-CP-sparse) dataflow. This approach accelerates the convolutional layers, but boasts added benefit of utilizing both redundant weights and activations to improve performance. However both approaches have been insufficiently tested on adequate databases, highlighting need for further research in improvements in network sparsity. Furthermore, computational costs can be reduced at deployment by using pruning [138]. Pruning is a means by which sparsity is created. It can involve setting weights below a given threshold to zero, before retraining to regain accuracy. This achieves a smaller, more efficient, yet accurate network. For example, [137] uses a three stage approach to reduce network size and computational cost by feature map pruning in each convolutional layer. Stage 1 involves training of a CNN using parameters that ensure feature map size is only modified in max pooling layers. The second stage involves utilization of a screening strategy that calculates discriminability values from feature maps – convolutional and feature maps with low discriminability magnitudes are pruned. This is followed by a third stage which involves piecewise pruning and retraining of each convolutional layer in the network. Often, DCNNs can be pruned significantly without affecting accuracy [137], within the range of approx. 20-80% [138].

The importance of sparsity, selectiveness and robustness was also emphasized by [142], which designed DeepID2+, improving upon DeepID2 by increasing the dimension of hidden representations (128 feature maps were used) and adding supervision to early convolutional layers, improving accuracy by 1.98% to achieve 98.70% accuracy on the LFW dataset. This study achieved an accuracy of 99.47% on LFW by combining 25 DeepID2+ networks. This study also reflected the Hebbian principle, as it noted that neural activations are moderately sparse – different identities activate different subsets of neurons, while identical identities in different images activate similar neurons. It was suggested that binary activation patterns are important in reducing computational cost, further speculating that higher layers are sensitive to global features rather than local variations which may result from occlusion. However, as shown by [11], the depth and width of the DCNN is significant to allow adequate learning. The researchers successfully trained one of the largest CNNs on ImageNet, achieving the best results ever recorded on the database at the time. The network comprised of 8 learned layers: 5 convolutional layers and 3 fully connected layers. It was noted that removing layers reduced performance. Overall, they constructed a highly optimized GPU implementation which prevented overfitting by using data augmentation and dropout and used minimal preprocessing – the only preprocessing involved was resizing of images. The network was trained on original RGB images using Relu optimization, rather than tanh, as it meant training time occurred six times faster and input normalization was not required. Other notable face recognition systems include [8] which presented a CNNN system, comprised of several DCNNs, designed to perform unconstrained face detection and preprocessing, automated verification and recognition. The face detection module employed the deep pyramidal deformable parts model proposed by [99] which has the ability to detect faces with varying sizes and poses in unconstrained conditions, combined with the architecture proposed by [11] to extract deep features. This system was quantitatively evaluated on a range of datasets included Labeled Faces in the Wild (LFW) and the JANUS CS2, the latter containing highly challenging images and videos. Thus it can be observed that increasing depth increases accuracy [136] but has significant computational costs [9] and energy consumption [138]. Increasing photo resolution also increases computational cost [9], which provides another area which may benefit from additional research, despite the natural improvements in results consequent of greater access to bigger datasets and faster GPUs [11].

In most DCNNs occlusion is naturally included in learned features alongside non-occluded features, reducing accuracy by corrupting extracted image features. [151] proposed to rectify issues caused by occlusion using MaskNet, a differentiable module, optimized by end to end training, which adaptively generates various feature map masks for a range of occluded faces, which can be incorporated into CNN architectures. It functions by assigning higher weights to occluded features activated by non-occluded features, and lower weights to those activated by occluded features, improving recognition in images affected by synthetic and natural occlusion. It is claimed that the computational cost of adding the module is minimal. Furthermore, [44] attempted to rectify blurring and issues arising from large standoffs and poor image quality, caused by unconstrained illumination and environmental degradation, by incorporating a sensitivity term into a DCNN cost function. The sensitivity term is comprised of the error divided by the activation function derivative, followed by minimization of the total error using the gradient descent method. This improves generalization and feature extraction by shifting the neural activations of the hidden layers to the middle high gradient area of the activation function. This method demonstrated effectiveness in day and night time images, and at varying standoff distances on the Long Distance Heterogeneous Face dataset [152], however it was only tested on a small, augmented dataset of images. One area of benefit is its effectiveness in reducing the need for intensive manual preprocessing of highly blurred images. Another early methodology which achieved competitive results without the benefit of large scale anotated datasets was proposed by [153] which used deep convolutional belief networks based on local convolutional restricted Boltzmann machines. Unsupervised representations were learned from unlabeled images, prior to being transferred to a recognition task by using a classification model such as support vector machine (SVM) and metric learning methods. FaceNet [67] also trained a DCNN with a very large database 100-200 million, using triplets of roughly aligned matching and non-matching face patches. [154] alternatively proposed handling pose variations by using several models, tailored for each major pose variation, resulting in greater accuracy. Other methods such as [48, 53] and [155] focused on establishing databases or improving existing databases as a means to improve learning via improved annotation, enlargement of databases, and handling of cross-pose factors using 3D morphable models.

## VIII. CONCLUSION

This survey presented a critical analysis of modern face recognition methodologies, developments and challenges. It also provided a comparative analysis of the available databases, and related benchmarks. It highlighted shortcomings of state of the art methods, and evaluated responses designed to address these limitations, emphasizing outstanding issues yet to be addressed. Despite drastic improvements in accuracy of representation due to the non-linearity of deep feature representations, we can confidently conclude that there is no known ideal facial feature that is sufficiently robust for face recognition in unconstrained environments. It must also be noted that solutions achieving state of the art accuracy are largely inhibited by their dependence on sophisticated GPUs and large databases, meaning there is still adequate need to focus research attention on more traditional handcrafted feature representations. Thus, the focus of future research must be on reducing the excessive computational cost of DCNNs, and their dependence on large, accurately annotated databases. Refinement of pruning methods, and minimization of training time is also an area requiring attention, as is network architecture, which would benefit from increased sparsity, and selectiveness.

## IX. CONFLICT OF INTEREST

This research is not affected by any conflicts of interest.

## X. FUNDING STATEMENT

This research is supported by an Australian Government Research Training Program (RTP) Scholarship.